\documentclass[10pt,twocolumn,letterpaper]{article}
\usepackage[left=0.75in,right=0.75in,top=0.8in,bottom=0.9in,columnsep=0.25in]{geometry}
\usepackage[utf8]{inputenc}
\usepackage[T1]{fontenc}
\usepackage{mathptmx}
\usepackage[scaled=0.92]{helvet}
\usepackage{amsmath,amssymb}
\usepackage{booktabs}
\usepackage{graphicx}
\usepackage{algorithm}
\usepackage[noend]{algpseudocode}
\usepackage{tikz}
\usetikzlibrary{arrows.meta,positioning}
\usepackage{microtype}

\DeclareUnicodeCharacter{1EBF}{\'{\^e}}
\DeclareUnicodeCharacter{1EC5}{\~{\^e}}
\DeclareUnicodeCharacter{0169}{\~u}

\makeatletter
\renewcommand\section{\@startsection{section}{1}{\z@}{1.1ex plus .3ex minus .2ex}{0.6ex plus .1ex}{\normalfont\normalsize\bfseries\scshape}}
\renewcommand\subsection{\@startsection{subsection}{2}{\z@}{0.9ex plus .2ex minus .1ex}{0.4ex plus .1ex}{\normalfont\normalsize\bfseries\itshape}}

\renewcommand\@seccntformat[1]{\csname the#1\endcsname.\hspace{0.5em}}
\makeatother

\algrenewcommand\algorithmicindent{1em}
\algrenewcommand\alglinenumber[1]{\scriptsize #1:}
\newcommand{\Tt}[1]{\texttt{#1}}
\newcommand{\tmat}{t}
\newcommand{\cmat}{c}

\title{\bfseries A Hybrid Exact--Heuristic Solver for the Integrated\\ Passenger and Parcel Transportation Problem\\[2pt]
\large SOICT 2026 Student Contest --- Technical Report}
\author{\begin{tabular}[t]{c}\textbf{Bùi Thế Dũng}\\ {\small\texttt{25023189@vnu.edu.vn}}\end{tabular}
\qquad\qquad
\begin{tabular}[t]{c}\textbf{Nguyễn Hà Thu}\\ {\small\texttt{25023407@vnu.edu.vn}}\end{tabular}\\[4pt]
\normalsize University of Engineering and Technology, Vietnam National University, Hanoi (UET-VNU)}
\date{}

\begin{document}
\maketitle
\vspace{-1.5em}

\noindent\textbf{\textit{Abstract}---}We describe the solver we submitted to the second round of the SOICT 2026 contest on the \emph{Integrated Passenger and Parcel Transportation} problem, a selective pickup-and-delivery routing problem with time windows, open routes and heterogeneous taxis. The solver is a single-file, standard-library Python program that combines (i) a constant-time insertion test based on latest-arrival slack, (ii) greedy insertion, local search and an Adaptive Large Neighborhood Search (ALNS) with simulated-annealing acceptance, and (iii) an exact route-selection engine that enumerates, for every taxi, the best feasible route for each request set and then solves the resulting set-packing problem by branch-and-bound strengthened with route-based large neighborhood search, Lagrangian bounds and reduced-cost fixing. The engine is chosen automatically from the instance size, and an independent verifier and a watchdog guarantee that a valid answer is always printed. On the official evaluation (38 scored instances, 5-minute limit) the solver obtained a total benefit of 487,237, finished every run within 277 s and used at most 279 MB; 28 of the 38 instances terminated well before the limit.


\section{Introduction}
A transportation company operates $K$ taxis that can carry passengers and parcels. Given $N$ passenger requests and $M$ parcel requests, each with a revenue, the task is to choose which requests to serve, assign them to taxis and order the stops so that revenue minus travel cost (the \emph{benefit}) is maximized, subject to time windows and parcel capacity \cite{dung2026}. The problem belongs to the family of pickup-and-delivery problems with time windows (PDPTW) \cite{parragh2008}, with three twists: requests may be rejected (a \emph{profit} or \emph{selective} routing problem \cite{feillet2005}), routes are open, and a passenger must be driven directly from pickup to drop-off while parcels may stay on board during such a trip.

Instance sizes are not bounded in the statement; in practice they range from a few requests to $250+250$ requests. No single method is best over this range: for small and medium instances a route-based exact method can prove optimality, while for the largest ones the number of feasible routes explodes and a metaheuristic is required. We combine both in one pipeline that switches automatically, together with engineering details (fast feasibility test, symmetry breaking, pool reduction, safety net) that make it robust under a strict time limit. The individual components are known (Section~\ref{sec:related}), and we do not claim algorithmic novelty.

\section{Related Work}\label{sec:related}
The problem is closely related to the Share-a-Ride Problem, in which people and parcels share taxis \cite{li2014}; an ALNS heuristic for that problem is given in \cite{li2016}. Our ALNS follows the framework of \cite{ropke2006} with Shaw removal \cite{shaw1998}. Exact algorithms for the PDPTW based on a set-partitioning-like formulation, dual bounds and the removal of routes whose reduced cost exceeds the optimality gap are described in \cite{baldacci2011}; our exact engine is a lighter relative of this idea that uses explicit route enumeration, branch-and-bound and Lagrangian bounds without column generation or cuts. Combining a metaheuristic with set partitioning over routes \cite{subramanian2013} and re-optimizing subsets of routes exactly \cite{queiroga2021} are likewise established matheuristic ideas. The contribution of this report is therefore the concrete integration for this contest problem under a 5-minute, single-file, standard-library constraint, and the empirical observations of Section~\ref{sec:eval}.

\section{Problem and Notation}
Taxi $k$ starts at depot $O_k$ at time $0$ with parcel capacity $Q_k$ and may stop anywhere. Passenger request $i$ has pickup $P_i$, drop-off $D_i$, revenue $T_i$, pickup window $[E_i,L_i]$ and service time $S_i$; after pickup the taxi drives straight to $D_i$ and no other request point may be visited in between. Parcel request $j$ has pickup and delivery points, windows $[E^p_j,L^p_j]$ and $[E^d_j,L^d_j]$, service time $S_j$, weight $w_j$ and revenue $T_j$; pickup and delivery are done by the same taxi, pickup first, and the load may never exceed $Q_k$. Arriving early means waiting until the window opens; arriving after $L$ is infeasible. With travel time $\tmat(u,v)$ and cost $\cmat(u,v)$ (both directed, non-negative, no triangle inequality assumed), the objective is
\begin{equation}
\max\; \sum_{i\in S_N}T_i+\sum_{j\in S_M}T_j-\sum_{k}\sum_{r=1}^{m_k}\cmat(v^k_{r-1},v^k_r).
\end{equation}
\textbf{Key observation.} The cost does not depend on time; time only enters through feasibility. Hence two insertions can be compared by their cost increase alone, and feasibility can be tested with slack information (Section~\ref{sec:slack}).

\section{Representation and Fast Feasibility}\label{sec:slack}
A solution stores, for each taxi, the list of its stops, and for each request the taxi serving it (or none). A route is always \emph{replaced} rather than edited in place, so snapshots for undo are cheap.

\textbf{Route evaluation.} For route $(v_0{=}O_k,v_1,\dots,v_n)$ a forward pass computes arrival $a_i=d_{i-1}+\tmat(v_{i-1},v_i)$, rejects if $a_i>L_{v_i}$, sets the departure $d_i=\max(a_i,E_{v_i})+S_{v_i}$, and tracks the load. A backward pass computes the latest feasible arrival
\begin{equation}
\ell_n=L_{v_n},\quad \ell_{i}=\min\bigl(L_{v_i},\ \ell_{i+1}-\tmat(v_i,v_{i+1})-S_{v_i}\bigr).
\end{equation}
Because the route is feasible, a block of new stops inserted between $v_i$ and $v_{i+1}$ and ending with departure time $d$ at node $z$ keeps the time windows of the tail satisfied \emph{iff} $d+\tmat(z,v_{i+1})\le \ell_{i+1}$, so every insertion position is tested in $O(1)$ \cite{savelsbergh1992}.

\textbf{Insertion scan.} For a request $r$ and taxi $k$, \Tt{scan} returns the cheapest feasible insertion and its cost increase. A passenger is inserted as the contiguous block $(P,D)$ and never directly after another passenger pickup. A parcel needs two positions; between them the delayed stops are re-simulated and the scan stops as soon as a window or the capacity is violated. Every accepted insertion is re-evaluated from scratch, so the stored solution is always feasible even if the scan were wrong. Results of \Tt{scan} are cached per (request, taxi) and invalidated when a route changes.

\section{Solver Pipeline}
\begin{figure}[t]
\centering
\begin{tikzpicture}[font=\scriptsize,node distance=3.2mm,>=Stealth,
 b/.style={draw,rounded corners=1pt,align=center,inner sep=2pt,minimum height=5.2mm}]
\node[b] (g) {Greedy\\insertion};
\node[b,right=of g] (l) {Local\\search};
\node[b,right=of l] (p) {Route\\enumeration\\(3 attempts)};
\node[b,above right=1mm and 6mm of p] (e) {Exact selection\\B\&B + LNS +\\Lagrange};
\node[b,below right=1mm and 6mm of p] (a) {ALNS\\+ polish};
\node[b,right=19mm of p] (v) {Verify\\\& print};
\draw[->] (g)--(l); \draw[->] (l)--(p);
\draw[->] (p.north east)|-(e.west) node[pos=.25,above,xshift=-2mm]{ok};
\draw[->] (p.south east)|-(a.west) node[pos=.25,below,xshift=-2mm]{fail};
\draw[->] (e.east)-|(v.north); \draw[->] (a.east)-|(v.south);
\end{tikzpicture}
\caption{Solver pipeline. The best verified solution found so far is always kept.}\label{fig:flow}
\end{figure}
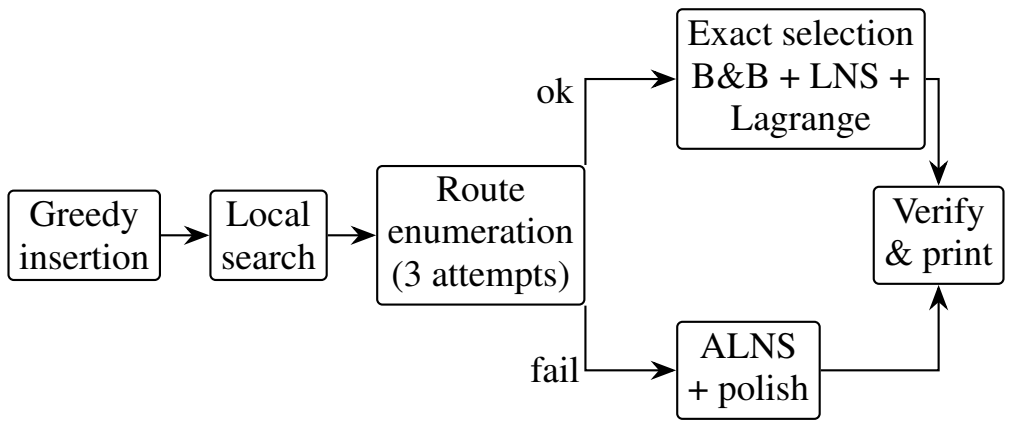

Figure~\ref{fig:flow} and Algorithm~\ref{alg:solve} show the control flow. A greedy solution and a local-search solution are obtained first and kept as safe answers. The solver then tries to enumerate the feasible routes of every taxi with three increasingly restrictive attempts: unrestricted routes (budget 10\% of the time limit), at most 4 requests per route (16\%), and at most 3 (12\%). If an attempt completes, the exact engine of Section~\ref{sec:exact} takes over and the heuristic engine is not used. If all three fail (too many states, routes or too slow), ALNS runs until the deadline. The time limit is $\mathrm{TL}=270$ s; a final polish of at most 8 s follows ALNS.

\begin{algorithm}[t]
\caption{\textsc{Solve}$(I)$}\label{alg:solve}
\begin{algorithmic}[1]
\State $S\gets$ \textsc{Greedy}$(I)$; $S\gets$ \textsc{LocalSearch}$(S)$; keep $S$
\State $A\gets$ \textsc{AliveRequests}$(I)$ \Comment{drop hopeless requests}
\For{$(\varphi,m)\in\{(.10,\infty),(.16,4),(.12,3)\}$}
  \State $\mathcal P\gets$ \textsc{BuildPool}$(I,A,m,\varphi\cdot\mathrm{TL})$
  \If{$\mathcal P\ne\bot$} \textbf{break}\EndIf
\EndFor
\If{$\mathcal P\ne\bot$}
  \If{$|\mathcal P|>40{,}000$} $\mathcal P\gets$ \textsc{ShrinkPools}$(\mathcal P)$ \EndIf
  \State \textsc{ExactSolve}$(\mathcal P)$ \Comment{Algorithm~\ref{alg:exact}}
\Else
  \State \textsc{ALNS}$(S)$; \textsc{LocalSearch}$(S)$ \Comment{Algorithm~\ref{alg:alns}}
\EndIf
\State print the best \emph{verified} solution
\end{algorithmic}
\end{algorithm}

\section{Heuristic Engine}
\textbf{Greedy and local search.} Greedy repeatedly inserts the (request, taxi) pair with the largest gain $T_r-\Delta\mathrm{cost}$ while the gain is positive, refreshing only the column of the modified taxi. Local search applies, with first improvement, \emph{relocate} (remove a request and re-insert it at its best place or drop it), \emph{swap} (exchange two requests between taxis) and \emph{replace} (swap a served request for an unserved one), undoing any move that does not improve the benefit.

\textbf{ALNS.} Algorithm~\ref{alg:alns} follows \cite{ropke2006}. \emph{Destroy} operators remove $q=\mathrm{rand}\{1,\dots,\max(2,\min(14,\lfloor0.3\,n_s\rfloor))\}$ served requests ($n_s$ is the number served): random; worst (rank by revenue minus the cost saved by removing the request, rank-biased); and Shaw removal \cite{shaw1998}, which removes requests related to a seed, with relatedness $\bigl(\tmat(P_s,P_r)+\tmat(D_s,D_r)\bigr)\tfrac{0.5}{\bar t}+|E_s-E_r|/t_{\max}$ ($\bar t$: mean travel time, $t_{\max}$: largest finite latest time), in two variants (served seed, or unserved seed that is re-offered to the pool). \emph{Repair} operators are greedy insertion and regret-2/regret-3 insertion, where the regret of a request is $\sum_{h=2}^{m}(g_1-g_h)$ over its sorted per-taxi gains. The repair pool also contains up to 8 random unserved requests, which gives rejected requests a chance to return. Candidates are accepted by simulated annealing \cite{kirkpatrick1983} with a temperature that decays exponentially with \emph{elapsed time} from $0.12\,\bar T$ to $0.004\,\bar T$ ($\bar T$ is the mean revenue), so the schedule does not depend on machine speed. Operator weights are updated every 60 iterations with reaction factor $0.25$ (scores 33/9/1 for new best/improving/accepted), and the search returns to the best solution after 4000 iterations without a new best.

\begin{algorithm}[t]
\caption{\textsc{ALNS}$(S,\mathrm{deadline})$}\label{alg:alns}
\begin{algorithmic}[1]
\State $S^*\gets S$
\While{time remains}
  \State $\tau\gets T_0(T_1/T_0)^{\text{elapsed}/\text{total}}$
  \State pick destroy $d$ and repair $\rho$ by roulette on weights
  \State $S'\gets S$; $(R,X)\gets d(S',q)$; $\rho(S',R\cup X\cup U_8)$
  \If{$f(S')>f(S^*)$} $S^*\gets S'$, $\sigma\gets 33$
  \ElsIf{$f(S')>f(S)$} $\sigma\gets 9$
  \ElsIf{$u<e^{(f(S')-f(S))/\tau}$, $u\sim U(0,1)$} $\sigma\gets 1$
  \Else{} $\sigma\gets 0$; undo; \textbf{continue}
  \EndIf
  \State $S\gets S'$; credit $\sigma$ to $d,\rho$; every 60 its. update weights
  \State \textbf{if} 4000 its. without new best \textbf{then} $S\gets S^*$
\EndWhile
\State \Return $S^*$
\end{algorithmic}
\end{algorithm}

\section{Exact Engine}\label{sec:exact}
\subsection{Preprocessing}
\textbf{Hopeless requests.} A Dijkstra pass over $\tmat$ from all depots at time 0 (ignoring service and waiting) gives lower bounds $a_v$ on the arrival time at any node. A request is dropped if $a_P>L_P$, $a_D>L_D$ or $w$ exceeds the largest capacity. \textbf{Tightened pickup deadline.} For a parcel, if the triangle inequality holds approximately on a random sample of triples, pickup must happen by $\min(L^p,\,L^d-S-\tmat(P,D))$. The check passed on all nine first-round practice instances (an exhaustive check of all triples on the five smaller ones and $1.5\cdot10^6$ random triples on the four larger ones found no violation exceeding the smallest service time), so the tightening was active on all of them. If a network violates the inequality only rarely, the sample may miss it and some feasible routes may be pruned; this can only lower the solution quality, because every printed solution is re-verified. \textbf{Identical taxis.} Taxis with equal capacity and equal time/cost rows from their depots are grouped; they share one route pool and the branch-and-bound breaks the symmetry among them.

\subsection{Route enumeration}
For each taxi a depth-first search (Algorithm~\ref{alg:pool}) extends a partial route by starting a new request (a passenger as one block, a parcel as its pickup) or by delivering a parcel on board. A complete route (non-empty, no parcel on board, positive profit) is stored under the \emph{set of requests it serves}; since revenue depends only on the set, the cheapest route for each set dominates all others, which collapses the number of stored routes. Candidates for each node are kept sorted by pickup deadline so that infeasible prefixes are skipped with a binary search. The search is aborted when it exceeds a state budget, the time budget, or $260{,}000$ stored routes.

\begin{algorithm}[t]
\caption{\textsc{BuildPool}$(I,A,m)$ for one taxi $k$}\label{alg:pool}
\begin{algorithmic}[1]
\Procedure{DFS}{$u,t,\sigma$}\Comment{$\sigma$: search state}
 \If{route is complete and rev$-$cost$>0$}
   \State keep it in $\mathrm{best}[\mathrm{used}]$ if more profitable
 \EndIf
 \If{$n<m$}
  \For{each unused $r\in A$ reachable in time}
   \State passenger: add $P_r,D_r$; parcel: add $P_r$ (if it fits)
   \State recurse; undo
  \EndFor
 \EndIf
 \For{each parcel $r$ on board, deliverable in time}
   \State append $D_r$; recurse; undo
 \EndFor
\EndProcedure
\State \textbf{abort} on state, time or route-count limit
\end{algorithmic}
\end{algorithm}

\subsection{Route selection}
Let $\Omega_k$ be the pool of taxi $k$, $\pi_\rho$ the profit of route $\rho$ and $a_{\rho i}\in\{0,1\}$ indicate that $\rho$ serves request $i$. The remaining problem is the set-packing model
\begin{align}
\max\ &\sum_k\sum_{\rho\in\Omega_k}\pi_\rho x_\rho\ \ \text{s.t.}\ \ \sum_{\rho\in\Omega_k}x_\rho\le1\ \ \forall k,\\
&\sum_k\sum_{\rho\in\Omega_k}a_{\rho i}x_\rho\le1\ \ \forall i,\qquad x_\rho\in\{0,1\}.\notag
\end{align}
\textbf{Branch-and-bound.} Routes of each pool are sorted by decreasing profit. At a node, the bound is the sum, over taxi groups, of the best still-compatible routes (one per remaining taxi), ignoring conflicts among them; the node is pruned if it cannot beat the incumbent. The next taxi to branch on is the one whose best compatible route is most profitable; a ``taxi unused'' branch removes the whole group, and within a group only routes of increasing index may be chosen (symmetry breaking). A node or time limit makes the search report that it was \emph{cut}; if it ends uncut, the incumbent is optimal for the pool.

\textbf{Route-based LNS.} Repeatedly, a random subset of $s$ taxis (initially 6) is released and re-optimized exactly by branch-and-bound while the others stay fixed; $s$ shrinks to at least 4 when the node limit is hit and grows when the sub-problem is solved quickly. This also builds the first incumbent.

\textbf{Lagrangian bound.} Relaxing the request constraints with multipliers $\lambda\ge0$ gives the valid bound
\begin{equation}
\mathrm{UB}(\lambda)=\sum_i\lambda_i+\sum_k\max\Bigl\{0,\ \max_{\rho\in\Omega_k}\bigl(\pi_\rho-\textstyle\sum_i a_{\rho i}\lambda_i\bigr)\Bigr\},
\end{equation}
minimized by subgradient steps $\lambda\leftarrow\max(0,\lambda-\theta\frac{\mathrm{UB}-\mathrm{LB}}{\|g\|^2}g)$ with $g_i=1-\sum a_{\rho i}x_\rho$ \cite{fisher1981}, with $\theta$ starting at 2 and multiplied by 0.6 after 10 non-improving iterations.

\textbf{Reduced-cost fixing.} If $\bar\pi_\rho=\pi_\rho-\sum_i a_{\rho i}\lambda_i$ and $b_k$ is the best reduced profit of taxi $k$ (at least 0), any solution using $\rho$ has value at most $\mathrm{UB}-b_k+\bar\pi_\rho$, so $\rho$ is deleted when this is below the incumbent LB. This keeps the pool exact while shrinking it.

\textbf{Large pools.} If the pool exceeds 40,000 routes, a heuristic reduction first removes dominated routes (a route whose request set contains that of a not-less-profitable route of the same taxi), runs LNS and the Lagrangian bound, and keeps only routes whose reduced profit is within $0.16\,\mathrm{LB}/K$ of the best of their taxi. This step may discard good routes, so optimality is then claimed only for the reduced pool.

\begin{algorithm}[t]
\caption{\textsc{ExactSolve}$(\mathcal P,\mathrm{deadline})$}\label{alg:exact}
\begin{algorithmic}[1]
\State run B\&B on $\mathcal P$ for $\min(10\,\mathrm{s},5\%$ of remaining time$)$
\If{not cut} \Return \Comment{optimal for $\mathcal P$} \EndIf
\State $\mathrm{chunk}\gets 6$ s
\While{time remains}
  \State route-LNS for chunk seconds; $\mathrm{LB}\gets$ incumbent value
  \State \textbf{if} first pass \textbf{then} compute $\lambda,\mathrm{UB}$ (15\% of remaining time)
  \State $\mathcal P'\gets$ reduced-cost fixing of $\mathcal P$ with $(\lambda,\mathrm{UB},\mathrm{LB})$
  \State B\&B on $\mathcal P'$ with base $\mathrm{LB}$, for chunk seconds
  \State \textbf{if} improved \textbf{then} adopt; \textbf{if} not cut \textbf{then return}
  \State $\mathrm{chunk}\gets\min(1.5\,\mathrm{chunk},60)$
\EndWhile
\end{algorithmic}
\end{algorithm}

\section{Safety and Implementation}
The program is one Python 3.8 file (about 1,100 lines, standard library only; main parameters in Table~\ref{tab:par}). Every improving solution is stored in a global holder. An \emph{independent verifier}, written separately from the evaluation used in the search, re-simulates each route (windows, direct passenger trips, pickup-before-delivery, capacity, no repeated node) and recomputes the benefit; only verified solutions are printed, falling back to the greedy solution. All loops check the deadline, and a \Tt{SIGALRM} watchdog set at 290 s prints the best verified solution and exits if some step overruns. The fixed random seed makes runs reproducible up to timing effects.

\begin{table}[t]
\centering\footnotesize
\caption{Main parameters of the submitted solver.}\label{tab:par}
\begin{tabular}{@{}ll@{}}
\toprule
Time limit / watchdog / polish & 270 s / 290 s / $\le 8$ s\\
Enumeration budgets & 10\%, 16\%, 12\% of 270 s\\
Route cap; state limits & 260,000; 6M (30M if restricted)\\
Pool reduction (limit, $\delta$) & 40,000 routes; $0.16\,\mathrm{LB}/K$\\
ALNS $q$ & $\mathrm{rand}\{1..\max(2,\min(14,\lfloor0.3n_s\rfloor))\}$\\
Temperature & $0.12\to0.004$ $\times$ mean revenue\\
Weights; restart to best & every 60 its., $0.25$; 4000 its.\\
Route-LNS size; node limit & 6 (4..$K$); 30,000\\
Subgradient & $\theta_0{=}2$, $\times0.6$/10 its., $\le1500$ its.\\
\bottomrule
\end{tabular}
\end{table}

\section{Evaluation}\label{sec:eval}
\textbf{Official results.} The second round had 39 test cases; one of them does not count towards the score (its benefit, 217, equals our result on the sample instance), leaving 38 scored instances. Table~\ref{tab:res} groups them by wall-clock time. Benefit ranges from 248 to 95,784, the total is \textbf{487,237}, the longest run took 276.9 s and the peak memory was 278.6 MB. Eighteen instances finish in under one second and 28 in under 61 s; these are consistent with the exact engine proving optimality over the enumerated (and possibly reduced) pool. The ten instances that use the whole budget carry 67.5\% of the total benefit; presumably these are the large instances where full enumeration fails or branch-and-bound cannot close the gap. On eight of the nine first-round practice instances (from $2{+}2{+}2$ up to $20{+}120{+}120$ taxis/passengers/parcels) the solver terminates in under 30 s on our development machine, for example with benefits 5,702 (10/47/48), 4,285 (10/50/50) and 13,279 (20/120/120).

\begin{table}[t]
\centering\small
\caption{Official evaluation, 38 scored instances, grouped by run time.}\label{tab:res}
\begin{tabular}{@{}lrrrrr@{}}
\toprule
Run time & \# & Benefit & Share & Max s & Max MB\\
\midrule
$\le 5$ s & 22 & 61,588 & 12.6\% & 4.0 & 184.5\\
5--61 s & 6 & 96,773 & 19.9\% & 60.9 & 250.8\\
$\ge 269$ s (limit) & 10 & 328,876 & 67.5\% & 276.9 & 278.6\\
\midrule
Total & 38 & 487,237 & 100\% & 276.9 & 278.6\\
\bottomrule
\end{tabular}
\end{table}

\textbf{A negative result.} We also evaluated a variant that, when full enumeration fails, solves a \emph{truncated} pool (at most 3 requests per route and only the most profitable routes of each taxi) with the same exact machinery instead of falling back to ALNS. On the same evaluation it scored 472,027 (a 3.1\% decrease): the score changed on eight instances and all eight became worse, by 1.5\% to 23.5\%; six of them ended between 51 s and 153 s before the time limit, presumably because the exact search finished on the truncated problem and the remaining time was left unused. Truncating routes and pools therefore loses more than the ALNS search does, which is why the submitted solver keeps ALNS as its fallback.

\textbf{Profiling the heuristic engine.} On synthetic $250{+}250$-request instances obtained by cloning nodes of a practice instance, insertion scans take about half of the ALNS time, and about 96\% of the scans recompute a (request, taxi, route content) triple already seen. Caching scan results by route \emph{content} instead of by version doubled the number of ALNS iterations per second ($2.2\times$) and improved the benefit by about 2\% at equal time. This change, and a preliminary prototype that re-optimizes clusters of 4--8 taxis exactly (enumeration plus branch-and-bound) starting from an ALNS solution, which improved a 60-s ALNS solution by 7\% to 14\% within 60 s on one synthetic instance (larger clusters gave larger gains), were \emph{not} part of the submitted solver; both numbers are preliminary and come from synthetic data.

\section{Conclusion}
The submitted solver combines a fast feasibility test, a robust ALNS and an exact route-selection engine with Lagrangian reasoning, selected automatically by instance size and protected by a verifier and a watchdog. Its weakest point is the group of large instances, which hold two thirds of the benefit and are solved heuristically; the results above suggest that faster insertion evaluation and exact re-optimization of small taxi clusters are the most promising ways to improve them. On networks that violate the triangle inequality substantially, the deadline tightening is switched off automatically by the sampled check; for networks with rare violations it should be disabled or the check made exhaustive.

\end{document}